\documentclass[sigconf]{acmart}
\usepackage{multirow}
\usepackage{algorithm}
\usepackage{algpseudocode}
\usepackage{float}
\usepackage{subfig}
\usepackage{amsmath}
\usepackage{booktabs}
\usepackage{makecell}
\usepackage{bm}

\let\L\relax
\let\dim\relax
\DeclareMathOperator{\L}{\mathcal{L}}
\DeclareMathOperator{\D}{\mathcal{D}}
\DeclareMathOperator{\E}{\mathbb{E}}
\DeclareMathOperator{\eem}{em}
\DeclareMathOperator{\graph}{graph}
\DeclareMathOperator{\class}{class}
\DeclareMathOperator{\dim}{dim}

\AtBeginDocument{%
  }

\setcopyright{acmcopyright}
\copyrightyear{2023}
\acmYear{2023}

\acmConference[CIKM'23]{The 32nd ACM International
Conference on Information and Knowledge Management}{}{}

\begin{document}

\title{VN-Solver: Vision-based Neural Solver for Combinatorial Optimization over Graphs}


\author{Mina Samizadeh}
\affiliation{%
 \institution{University of Delaware}
 \city{Newark}
 \state{Delaware}
 \country{USA}
 \postcode{19711}
}
\email{minasmz@udel.edu}

\author{Guangmo Tong}
\affiliation{%
 \institution{University of Delaware}
 \city{Newark}
 \state{Delaware}
 \country{USA}
 }
\email{amotong@udel.edu}


\begin{abstract}
Data-driven approaches have been proven effective in solving combinatorial optimization problems over graphs such as the traveling salesman problems and the vehicle routing problem. The rationale behind such methods is that the input instances may follow distributions with salient patterns that can be leveraged to overcome the worst-case computational hardness. For optimization problems over graphs, the common practice of neural combinatorial solvers consumes the inputs in the form of adjacency matrices. In this paper, we explore a vision-based method that is conceptually novel: can neural models solve graph optimization problems by \textit{taking a look at the graph pattern}? Our results suggest that the performance of such vision-based methods is not only non-trivial but also comparable to the state-of-the-art matrix-based methods, which opens a new avenue for developing data-driven optimization solvers.

\end{abstract}





\maketitle

\section{Introduction}
Combinatorial optimization problems are not only of great theoretical interest but also central to various application domains \cite{korte2011combinatorial}. Traditional approaches suffer from their forbidding execution time and the need for hand-crafted algorithmic rules \cite{boussaid2013survey}. Recent years have witnessed a surge in the development of neural combinatorial solvers, aiming to efficiently solve combinatorial optimization problems through machine learning techniques \cite{bengio2021machine}. The premise behind such methodologies is that the input-solution pairs in practical applications often exhibit certain data-dependent distributions, which, once successfully learned, can be leveraged to circumvent the worst-case NP-hardness. As the inherent structure of many problems is often relational \cite{vesselinova2020learning}, it is of paramount interest to examine the potential of neural solvers for addressing optimization problems on graphs. We will particularly focus on deterministic combinatorial optimization problems in this paper. 

\begin{figure}[t]
\centering
\subfloat[]{\label{fig: simple_hamil}\includegraphics[width=0.11\textwidth]{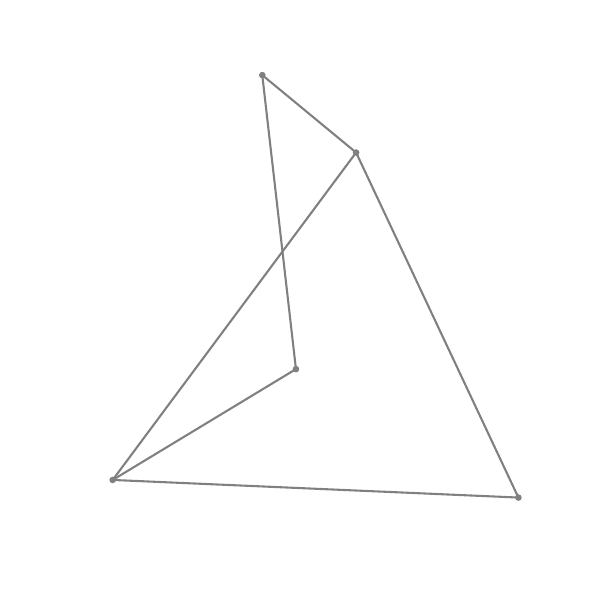}}\hspace{0mm}
\subfloat[]{\label{fig: simple_nonhamil}\includegraphics[width=0.11\textwidth]{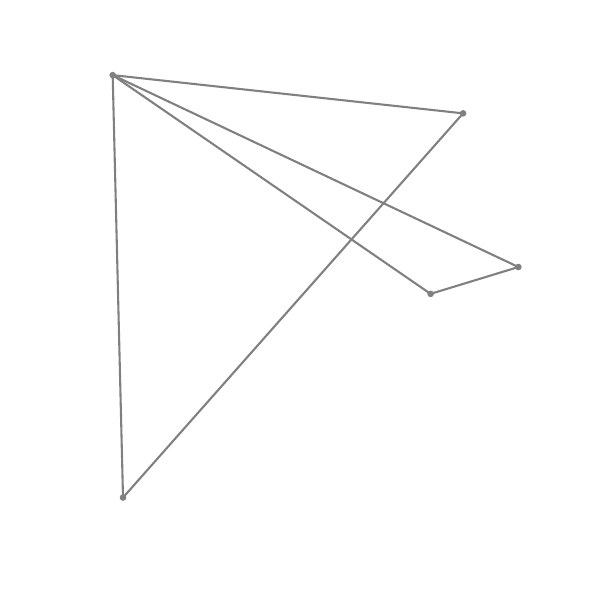}}\hspace{0mm}
\subfloat[]{\label{fig: complex}\includegraphics[width=0.11\textwidth]{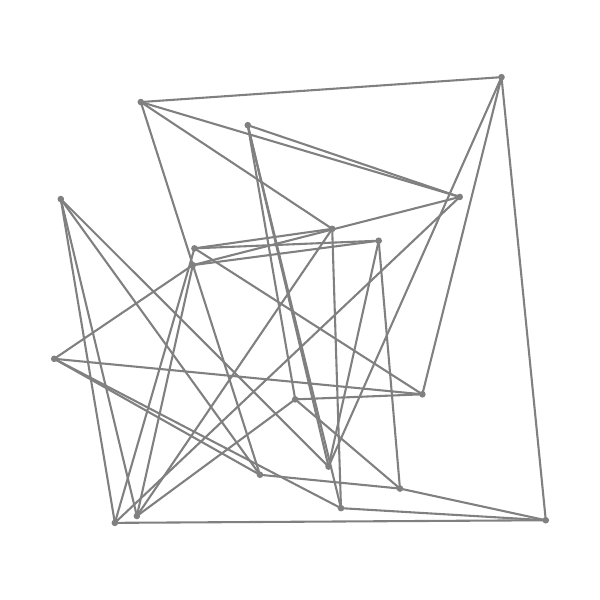}}\hspace{0mm}
\subfloat[]{\label{fig: complex_2}\includegraphics[width=0.11\textwidth]{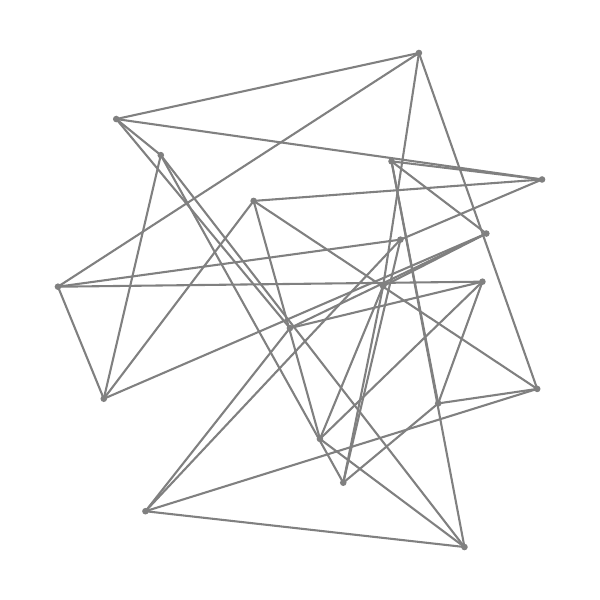}}
\caption{\small (a): a simple Hamiltonian graph; (b): a simple non-Hamiltonian graph; (c) and (d): visualizations that are relatively complicated.}
\label{fig: pipeline}
\vspace{-5mm}
\end{figure}

The state-of-the-art neural solvers are largely matrix-based, i.e., reasoning over the adjacency matrix by using deep neural networks \cite{cappart2021combinatorial}. In direct contrast to the matrix-based methods, this paper explores the so-called vision-based methods that consume graph visualizations as inputs. Such methods are desired when the input of the problem is given as images (e.g., maps of the traveling salesman problem \cite{perez2013automated}), as it does not require us to convert the images to matrices. More importantly,  we are motivated by the fact that for humans, visualizations can be much more intuitive than the adjacency matrix for certain instances. Taking the Hamiltonian cycle problem as an example, let us consider Figs. \ref{fig: simple_hamil} and \ref{fig: simple_nonhamil} with the following two adjacency matrices
\begin{align*}
A_a=\begin{bmatrix}
0 & 1 & 0 & 0 & 1\\
1 & 0 & 1 & 1 & 0\\
0 & 1 & 0 & 1 & 0\\
0 & 1 & 1 & 0 & 1\\
1 & 0 & 0 & 1 & 0
\end{bmatrix} \text{~~~and~~~} 
A_b=\begin{bmatrix}
0 & 1 & 1 & 1 & 1\\
1 & 0 & 0 & 0 & 1\\
1 & 0 & 0 & 1 & 0\\
1 & 0 & 1 & 0 & 0\\
1 & 1 & 0 & 0 & 0
\end{bmatrix}.
\end{align*}
It is obvious at a glance that Fig. \ref{fig: simple_hamil} is Hamiltonian while Fig. \ref{fig: simple_nonhamil} is not, but it would take arguably more time to figure out the same results if we were checking the adjacency matrices. There of course exist instances that cannot be easily recognized by humans -- for example, Figs. \ref{fig: complex} and \ref{fig: complex_2}, which are Hamiltonian and in fact correspond to the same graph. However, since deep learning methods have demonstrated visual perception ability better than humans \cite{szegedy2017inception,zou2023object}, it is possible that instances like Fig. \ref{fig: complex} can be successfully handled by image classifiers like ResNet \cite{targ2016resnet} or VIT \cite{yuan2021tokens}, without consulting the adjacency matrices. In this paper, we explore the feasibility of such methods.

\begin{figure*}[t!]
\centering
\subfloat[]{\label{fig: c-1-1-05-01-g}\includegraphics[width=0.10\textwidth]{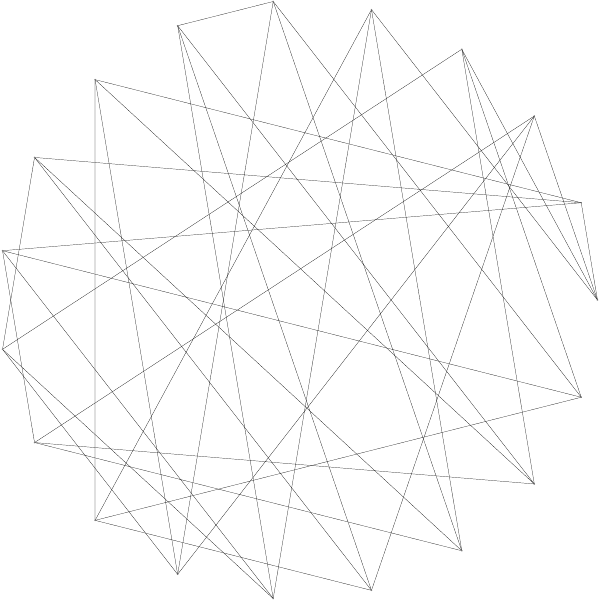}}\hspace{3mm}
\subfloat[]{\label{fig: c-1-05-05-01-g}\includegraphics[width=0.10\textwidth]{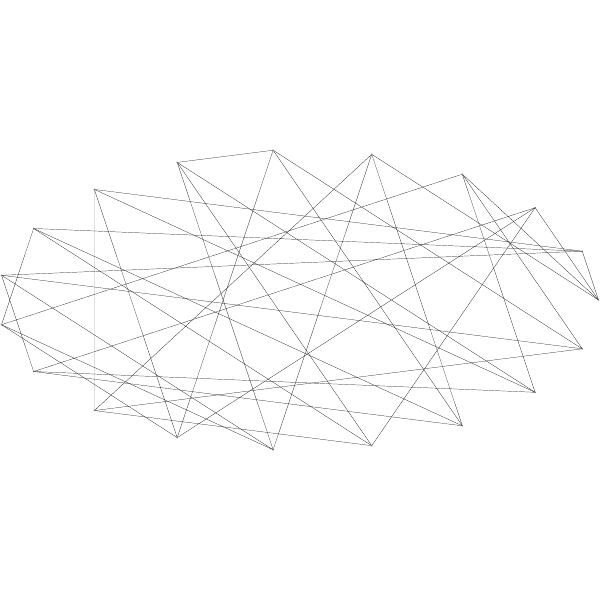}}\hspace{3mm}
\subfloat[]{\label{fig: c-1-01-05-01-g}\includegraphics[width=0.10\textwidth]{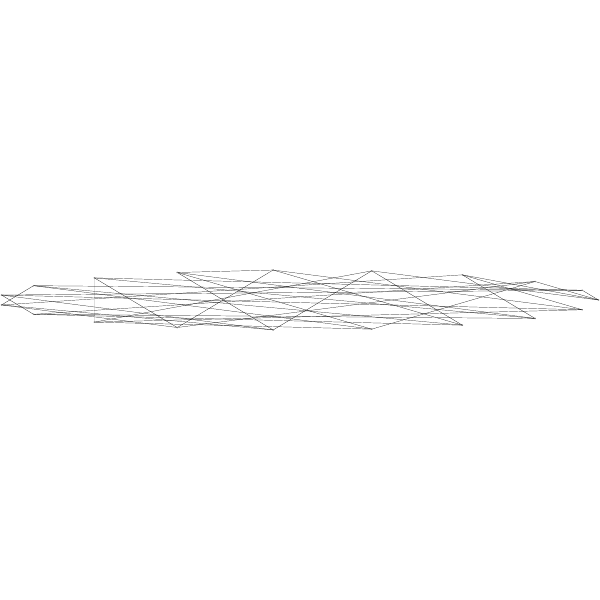}}\hspace{3mm}
\subfloat[]{\label{fig: c-1-001-05-01-g}\includegraphics[width=0.10\textwidth]{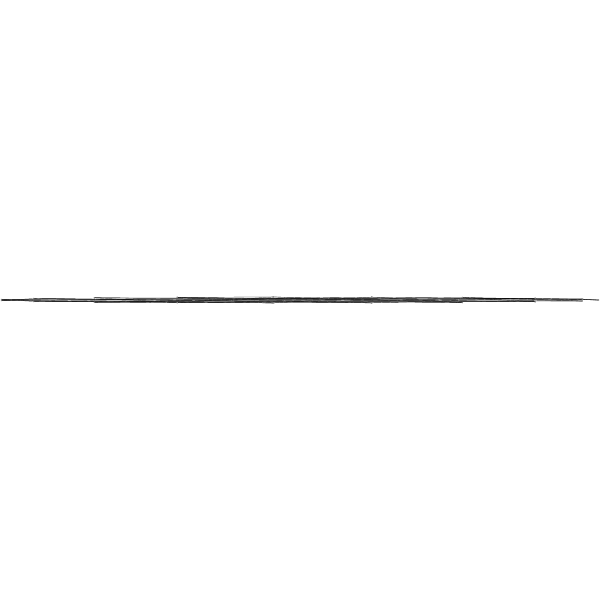}}\hspace{10mm}
\subfloat[]{\label{fig: s-01-05-01-g}\includegraphics[width=0.10\textwidth]{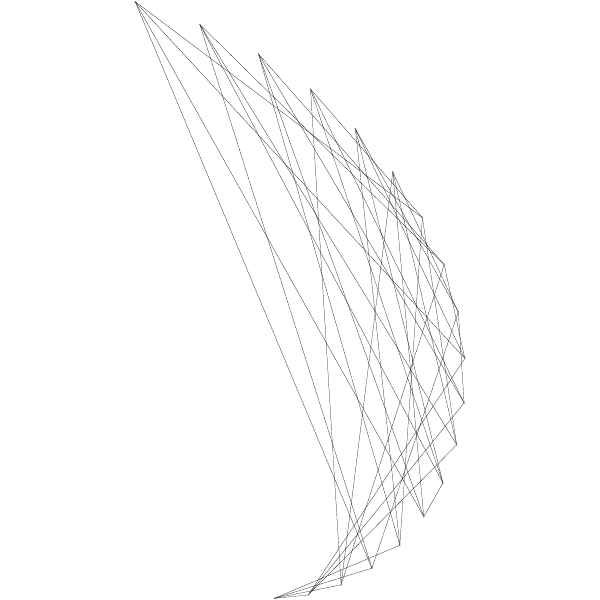}}\hspace{3mm}
\subfloat[]{\label{fig: s-03-05-01-g}\includegraphics[width=0.10\textwidth]{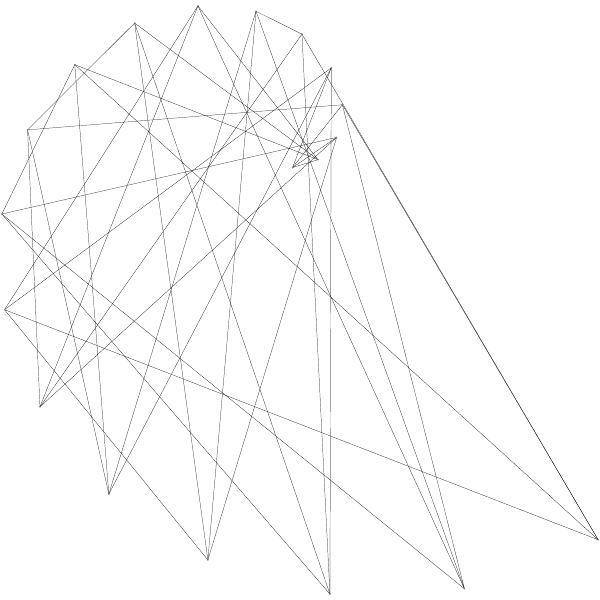}}\hspace{3mm}
\subfloat[]{\label{fig: s-05-05-01-g}\includegraphics[width=0.10\textwidth]{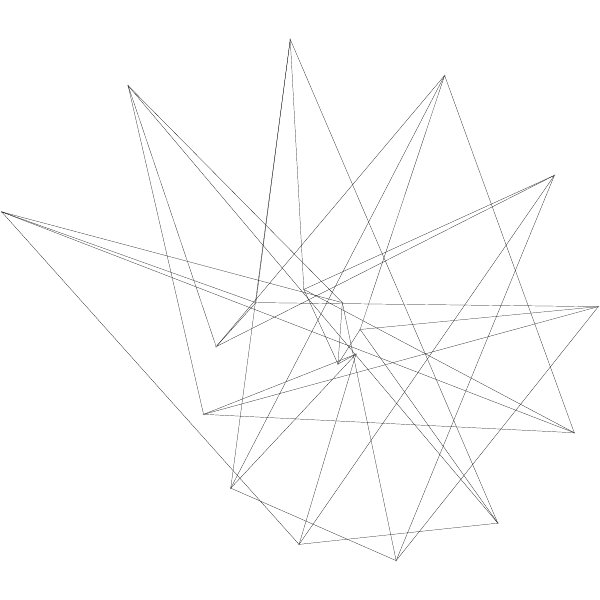}}\hspace{3mm}
\subfloat[]{\label{fig: s-07-05-01-g}\includegraphics[width=0.10\textwidth]{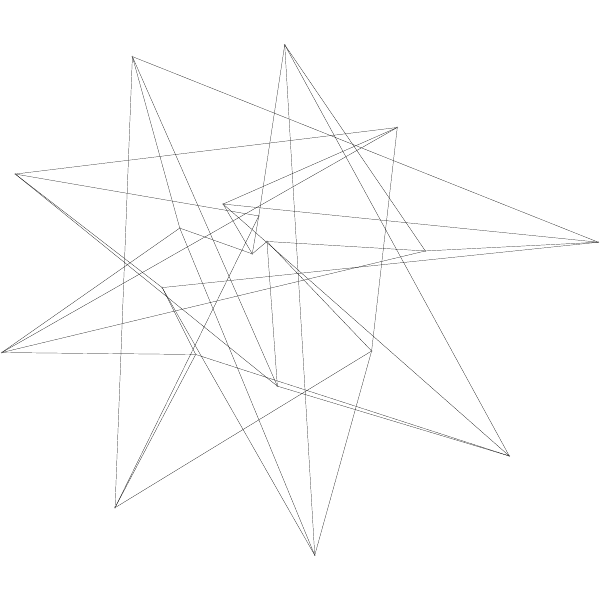}}

\subfloat[]{\label{fig: r-05-01-1}\includegraphics[width=0.10\textwidth]{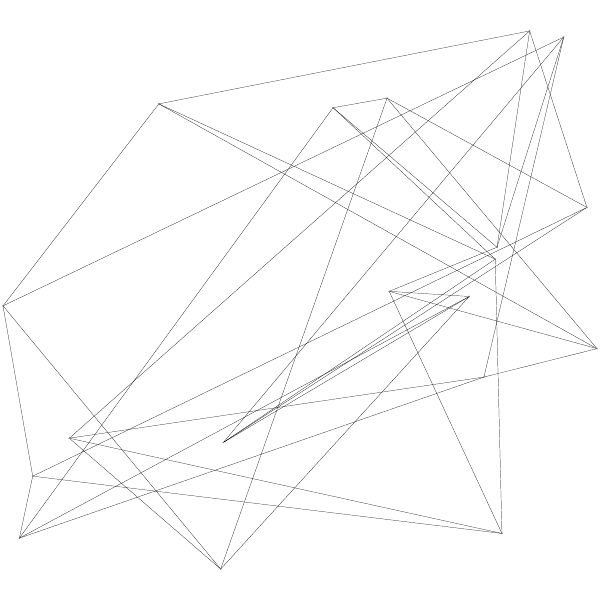}}\hspace{3mm}
\subfloat[]{\label{fig: r-05-01-2}\includegraphics[width=0.10\textwidth]{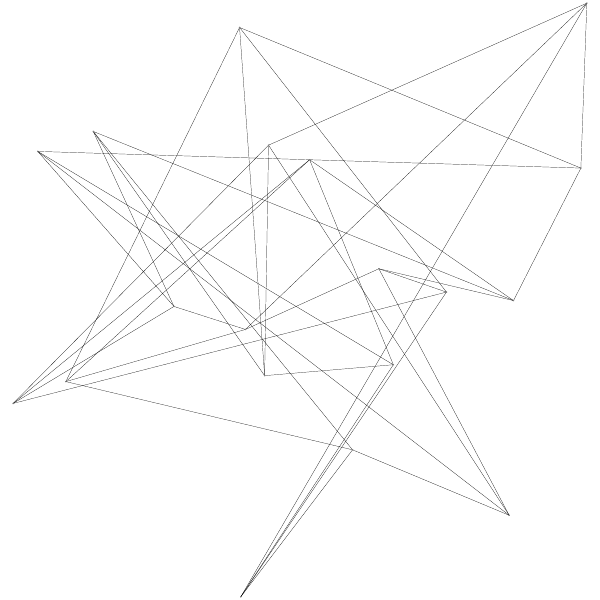}}\hspace{3mm}
\subfloat[]{\label{fig: r-05-01-3}\includegraphics[width=0.10\textwidth]{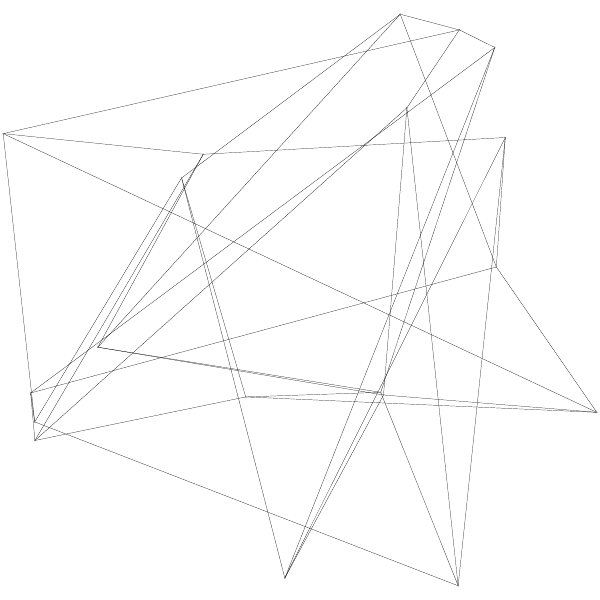}}\hspace{3mm}
\subfloat[]{\label{fig: r-05-01-4}\includegraphics[width=0.10\textwidth]{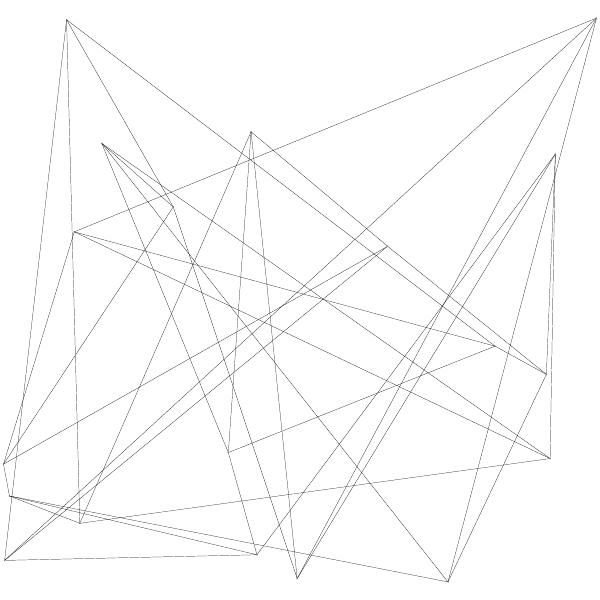}}\hspace{10mm}
\subfloat[]{\label{fig: c-1-1-0-01-g}\includegraphics[width=0.10\textwidth]{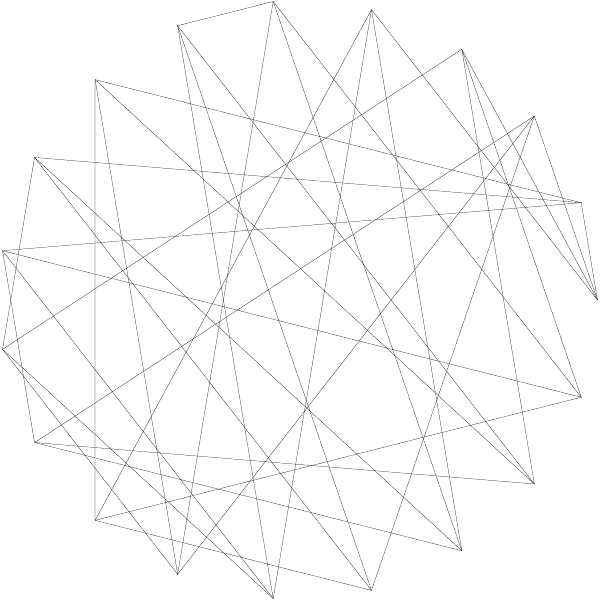}}\hspace{3mm}
\subfloat[]{\label{fig: c-1-1-200-1-g}\includegraphics[width=0.10\textwidth]{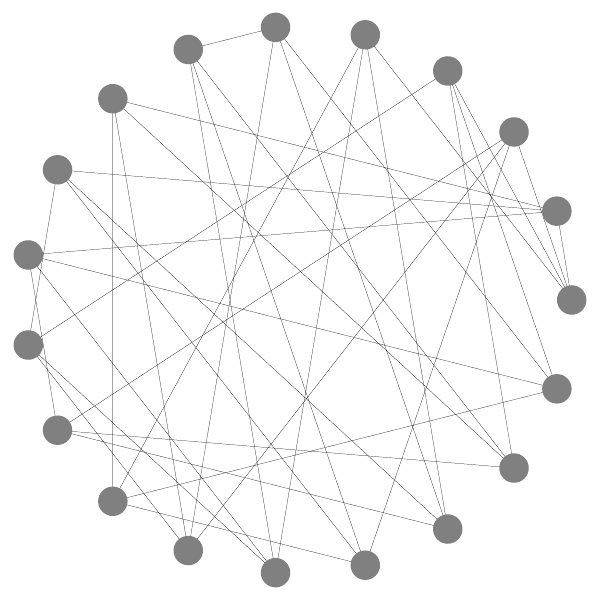}}\hspace{3mm}
\subfloat[]{\label{fig: c-1-1-300-10-g}\includegraphics[width=0.10\textwidth]{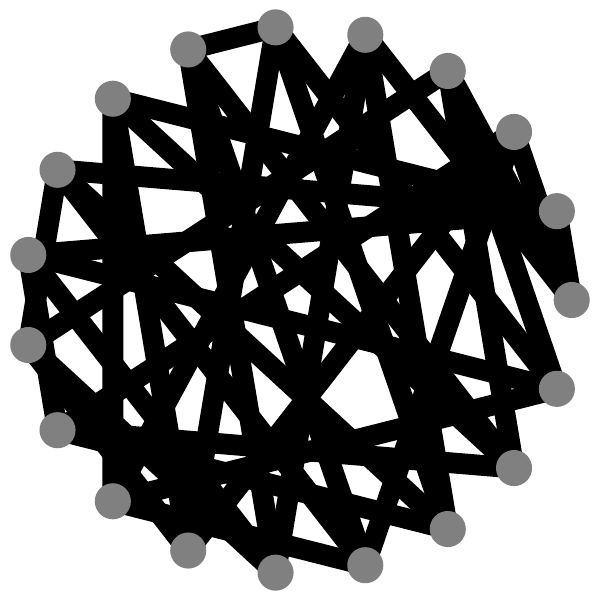}}\hspace{3mm}
\subfloat[]{\label{fig: c-1-1-100-100-g}\includegraphics[width=0.10\textwidth]{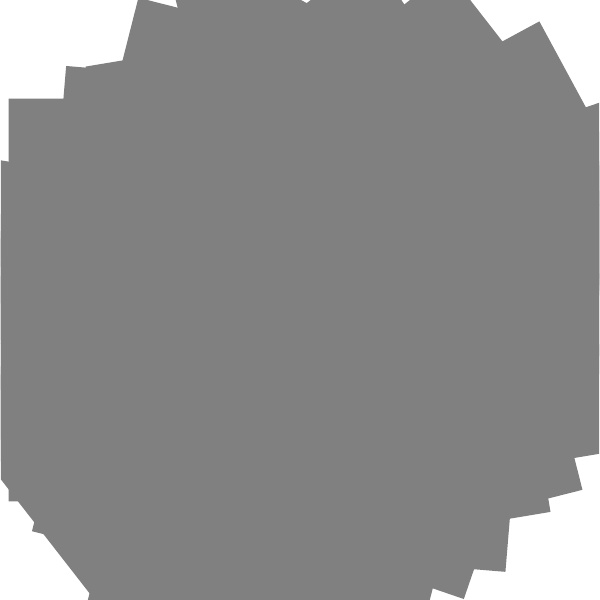}}

\subfloat[]{\label{fig: c-1-1-300-10-u}\includegraphics[width=0.10\textwidth]{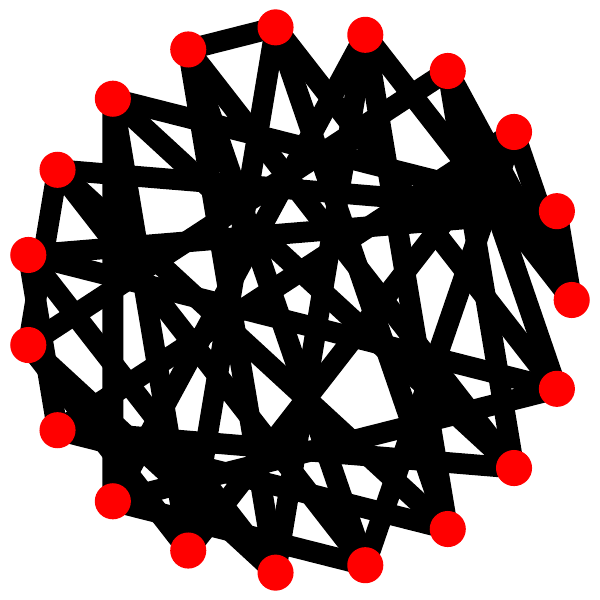}}\hspace{3mm}
\subfloat[]{\label{fig: s-03-300-10-u}\includegraphics[width=0.10\textwidth]{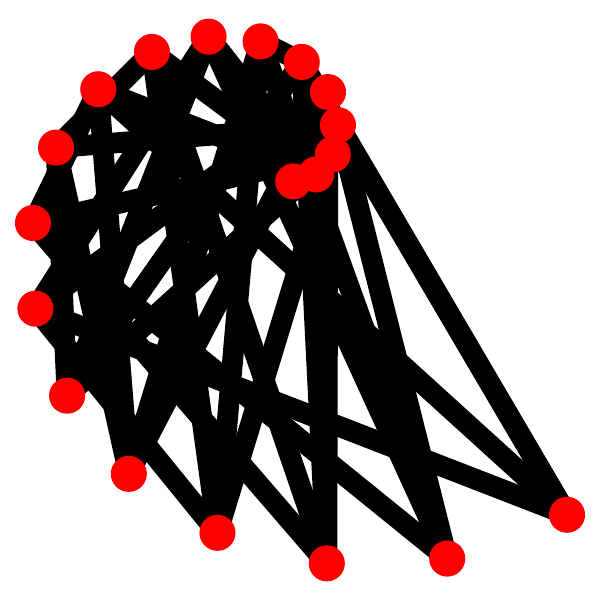}}\hspace{3mm}
\subfloat[]{\label{fig: r-300-10-u-1}\includegraphics[width=0.10\textwidth]{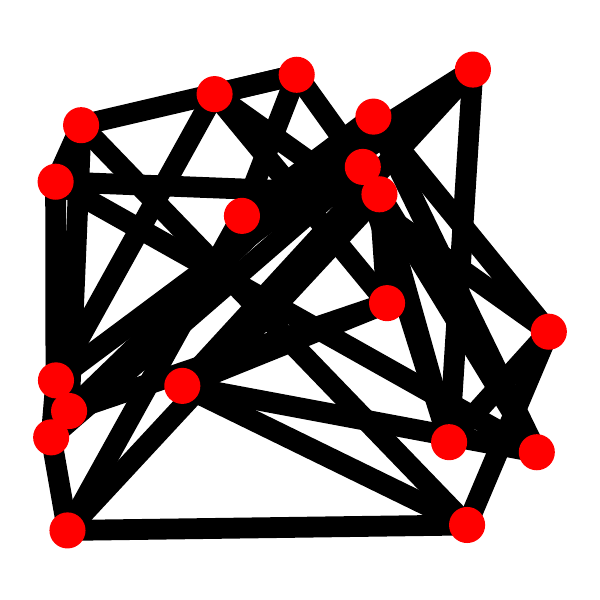}}\hspace{3mm}
\subfloat[]{\label{fig: r-300-10-u-2}\includegraphics[width=0.10\textwidth]{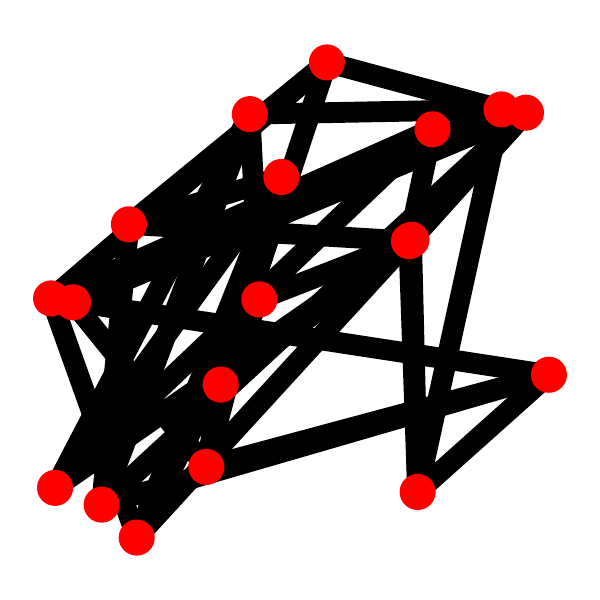}}\hspace{10mm}
\subfloat[]{\label{fig: c-1-1-300-10-r}\includegraphics[width=0.10\textwidth]{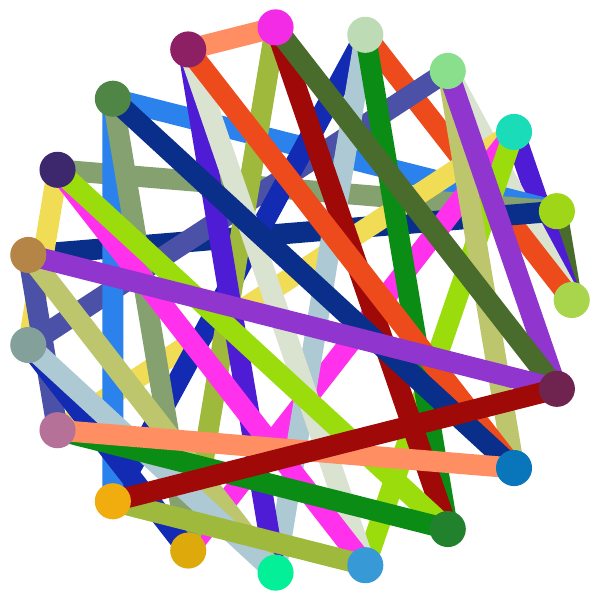}}\hspace{3mm}
\subfloat[]{\label{fig: s-03-300-10-r}\includegraphics[width=0.10\textwidth]{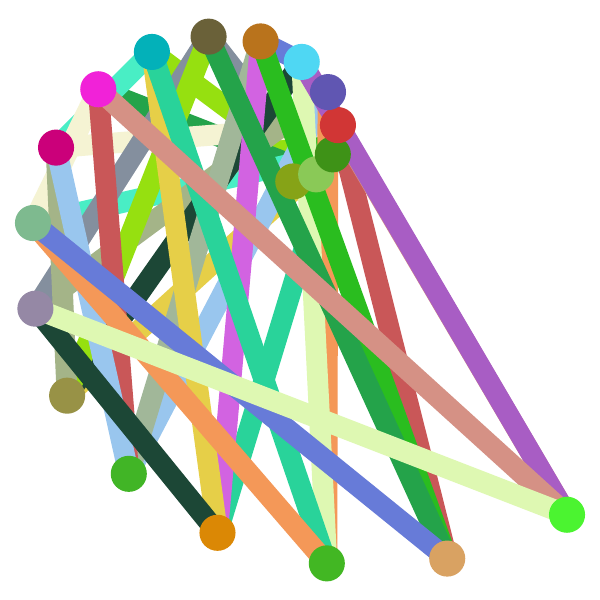}}\hspace{3mm}
\subfloat[]{\label{fig: r-300-10-r-1}\includegraphics[width=0.10\textwidth]{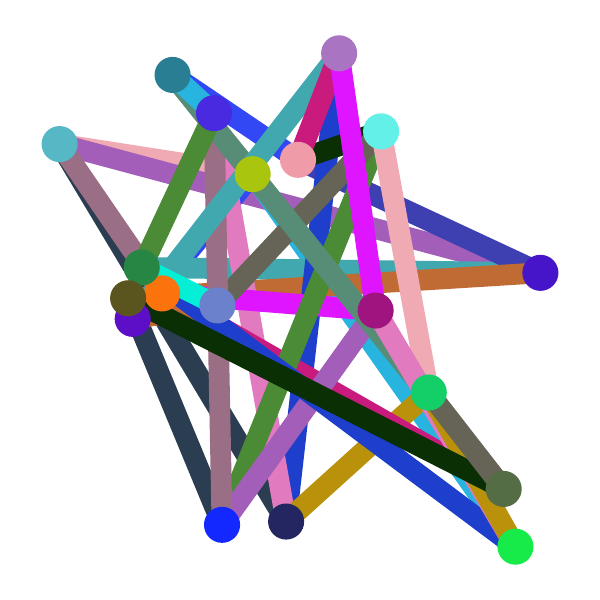}}\hspace{3mm}
\subfloat[]{\label{fig: r-300-10-r-2}\includegraphics[width=0.10\textwidth]{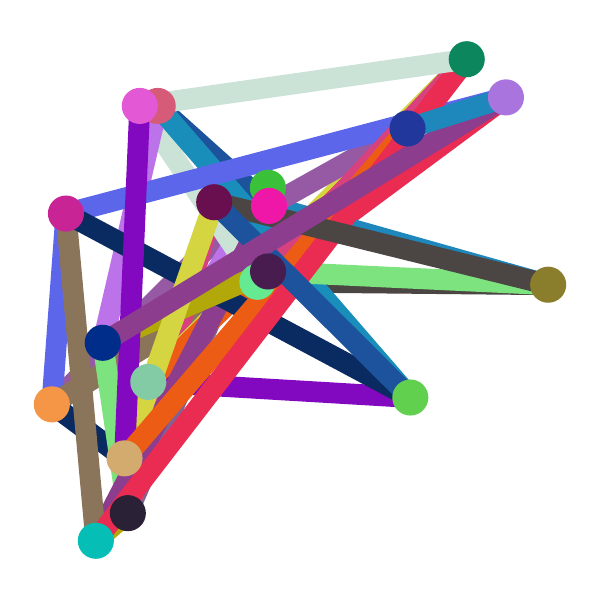}}\hspace{0mm}

\caption{\small Visualizations of different layouts, node sizes, edge thicknesses, and coloring schemes.
}
\label{fig: examples}
\vspace{-3mm}
\end{figure*}

\textbf{Contributions.} We present a conceptually simple framework called the vision-based method for solving deterministic combinatorial optimization problems over graphs. The proposed framework consists of three modules: \textit{graph embedding}, \textit{image generation}, and \textit{image classification}. The first module decides how to embed a graph in Euclidean space; the second module visualizes the embedding as an image of pixels; the third module decides if the input is a true instance. We adopt ResNet for image classification and discuss possible choices for graph embedding and image generation. Our experiments promisingly show that vision-based methods are effective with statistical significance and no less powerful than the state-of-the-art matrix-based methods for the Hamiltonian cycle problem. Our experiments also provide insights into the role of the visualization modules in determining the framework's efficacy. For example, we observe that structured visualizations are evidently better than random visualizations.


\section{VN-Solver: a Vision-based Method}
A deterministic graph optimization problem is specified by a function $F$ that maps each graph $G=(V, E)$ to true or false. For example, in planarity testing \cite{hopcroft1974efficient}, $F(G)=1$ if and only if $G$ is a planar graph. Some of such problems can be readily solved (e.g., connectivity testing \cite{even1975network}), while others are computationally hard under common complexity assumptions (e.g., Hamiltonian cycle or path \cite{rahman2005hamiltonian}). From the perspective of statistical learning, a neural solver seeks to learn $F$ using empirical evidence, i.e., pairs of $(G, F(G))$, with the goal of minimizing the generalization error
\begin{align}
    \L(\tilde{F}, \D)= \E_{G \sim \D} \big[ l(F(G), \tilde{F}(G)) \big],
\end{align}
where $\tilde{F}$ is the learned function, $\D$ is an unknown distribution over graphs, and, $l$ is a certain loss function. 

Matrix-based methods take the adjacency matrix $A_G$ of $G$ as the input and often process $A_G$ through various neural architectures where the last layer gives a distribution over $\{0,1\}$ suggesting if the input graph is a true instance. Our vision-based method is conceptually different and consists of three simple steps: 
\begin{align*}
    \text{\textbf{graph embedding}:~}& \hspace{5mm} F_{\eem} : V \rightarrow \mathbb{R}^2\\
    \text{\textbf{image generation}:~}&  \hspace{5mm} F_{\graph} : \mathbb{R}^{2 \times |V|} \rightarrow \mathbb{R}^{3 \times \dim_x \times \dim_y}\\
    \text{\textbf{image classification}:~}& \hspace{5mm} F_{\class} : \mathbb{R}^{3 \times \dim_x \times \dim_y} \rightarrow \{0, 1\}.
\end{align*}
$F_{\eem}$ maps each node $v$ to a point $(v_x,v_y) \in \mathbb{R}^2$ in the 2D Euclidean space, and each edge $(u,v)$ corresponds to the line segment with edge points $(u_x,u_y)$ and $(v_x,v_y)$. Based on the embedding (i.e., a collection of points in the 2D space), $F_{\graph}$ generates an image of size $\dim_x \times \dim_y$ in RGB pixels. Finally, a classifier $F_{\class}$ decides if the input is a true instance. We denote such a vision-based neural solver as VN-Solver. The recent advance in computer vision has provided us with a great collection of tools for image classification, and we will leverage them for the third step. In particular, we adopt the celebrated ResNet model \cite{he2016deep} in our experiments. While image classification is not a burden, the methods for graph embedding and image generation are open to different design principles.

For graph embedding, one can imagine that some methods will definitely not work -- for example, putting all the nodes on a straight line will make the edges not recognizable. We primarily focus on two principled methods: circular layout \cite{bhavsar2022graph} and spiral layout \cite{carlis1998interactive}. The circular layout puts the nodes uniformly on an eclipse $x^2/a+y^2/b=1$,
where the ratio between $a$ and $b$ controls the shape. Figs. \ref{fig: c-1-1-05-01-g}-\ref{fig: c-1-001-05-01-g} are examples with different ratios, where we can sense that an extreme ratio (e.g., Fig. \ref{fig: c-1-001-05-01-g}) may not be useful for recognizing graph properties (which we verify in experiments). For each $v_i \in V=\{v_1,...,v_n\}$, the spiral layout, dated back to the spiral of Archimedes \cite{czwalina1922spiralen}, assigns its coordinate as 
$(i\cdot \cos(i \cdot r), i\cdot \sin(i \cdot r)),$
where $r \in \mathbb{R}$ is the offset parameter. Examples with different offsets can be found in Figs. \ref{fig: s-01-05-01-g}-\ref{fig: s-07-05-01-g}. One can even use random embeddings where the coordinates of the nodes are uniformly random within, for example, $[0,1]\times [0,1]$ -- Figs. \ref{fig: r-05-01-1}-\ref{fig: r-05-01-4}.

In generating an image for a given embedding, we plot nodes as solid circles and edges as solid line segments. The size is $224$ by $224$ in pixels, which is aligned with ResNet \cite{he2016deep}. The primary factors we consider are the size of the node circles and the thickness of the edges segments (Figs. \ref{fig: c-1-1-0-01-g}-\ref{fig: c-1-1-100-100-g}). In addition to gray images (e.g., Figs. \ref{fig: c-1-1-05-01-g}-\ref{fig: c-1-1-100-100-g}), the images may be colored in various ways. For example, in the \textit{uniform-color} scheme, the nodes are in one color and the edges are in another color (e.g., Figs. \ref{fig: c-1-1-300-10-u}-\ref{fig: r-300-10-u-2}), while in the \textit{random-color} scheme, we randomly pick one color for each entity (i.e., node or edge) -- Figs. \ref{fig: c-1-1-300-10-r}-\ref{fig: r-300-10-r-2}. While human perception may not be very sensitive to these factors, images with different node circle sizes can be very different from each other from the perspective of machine learning models. In fact, all the visualizations in Fig. \ref{fig: examples} are associated with the same underlying graph. It would be interesting if one can observe that some of them are more informative than others in deciding certain graph properties.

\begin{table*}[pt!]
\renewcommand{\arraystretch}{1.2} 

\caption{\small Main results. Top F1 scores are highlighted. The sample size for training is from $\{100,200,1000\}$}
\vspace{-3mm}
\label{table: results_all}
\centering
{\small
\begin{tabular}{@{}  l@{\hspace{3mm}}  l@{\hspace{3mm}}  l@{\hspace{5mm}}  c @{\hspace{1mm}} c @{\hspace{1mm}} c @{\hspace{5mm}} c @{\hspace{1mm}} c @{\hspace{1mm}}c @{\hspace{5mm}} c @{\hspace{1mm}} c @{\hspace{1mm}} c @{}}
\toprule
\multicolumn{3}{c}{\multirow{2}{*}{\textbf{Dataset: Small}}} & \multicolumn{3}{c}{100}   &  \multicolumn{3}{c}{200} &  \multicolumn{3}{c}{1000} \\

&   & & AUC   & Accuracy & F1 & AUC   & Accuracy & F1 & AUC   & Accuracy & F1\\
\midrule
\multirow{7}{*}{VN-Solver}     & \multirow{3}{*}{Gray}  &  Circular & $0.55 \pm 0.06$ & $0.45 \pm 0.03$ & $0.62 \pm 0.03$  & $0.43 \pm 0.02$ & $0.44 \pm 0.02$ & $0.61 \pm 0.02$ & $0.86 \pm 0.09$ & $0.79 \pm 0.09$ & $\bm{0.78} \pm 0.08$\\
    &   &  Spiral & $0.59 \pm 0.15$ & $0.52 \pm 0.13$ & $0.63 \pm 0.04$ & ${0.65} \pm 0.14$ & $0.54 \pm 0.15$ & $0.65 \pm 0.06$ & $0.84 \pm 0.02$ & $0.78 \pm 0.02$ & $0.76 \pm 0.02$\\
     &   &  Random & $0.50 \pm 0.01$ & $0.49 \pm 0.06$ & $0.50 \pm 0.28$ & $0.51 \pm 0.02$ & $0.44 \pm 0.02$ & $0.61 \pm 0.02$ & $0.52 \pm 0.05$ & $0.49 \pm 0.07$ & $0.37 \pm 0.34$\\
         \cmidrule{2-12}
      & \multirow{3}{*}{\makecell{Uniform \\ color}}  &  Circular &   $0.69 \pm 0.13$ & $0.61 \pm 0.09$ & $\bm{0.63} \pm 0.09$ & $0.75 \pm 0.14$ & $0.68 \pm 0.12$ & $\bm{0.69} \pm 0.04$ & $0.93 \pm 0.02$ & $0.85 \pm 0.02$ & $\bm{0.83} \pm 0.03$\\
    &   &  Spiral &  $0.70 \pm 0.07$ & $0.62 \pm 0.05$ & $\bm{0.65} \pm 0.05$ & $0.78 \pm 0.06$ & $0.71 \pm 0.06$ & $\bm{0.72} \pm 0.05$ & $0.86 \pm 0.04$ & $0.78 \pm 0.06$ & $0.76 \pm 0.07$ \\
     &   &  Random & $0.52 \pm 0.03$ & $0.45 \pm 0.0$ & $0.62 \pm 0.00$ & $0.51 \pm 0.02$ & $0.43 \pm 0.00$ & $0.60 \pm 0.00$ & $0.51 \pm 0.02$ & $0.48 \pm 0.00 $ & $0.65 \pm 0.00$ \\
    \cmidrule{2-12}
          & \multirow{2}{*}{\makecell{Random \\ color}}  &  Circular & $0.63 \pm 0.09$ & $0.54 \pm 0.11$ & $0.61 \pm 0.03$ & $0.65 \pm 0.11$ & $0.56 \pm 0.11$ & $0.64 \pm 0.04$ & $0.9 \pm 0.02$ & $0.83 \pm 0.03$ & $\bm{0.81} \pm 0.02$ \\
          &  &  Spiral & $0.73 \pm 0.06$ & $0.62 \pm 0.08$ & $\bm{0.64} \pm 0.04$ & $0.64 \pm 0.15$ & $0.6 \pm 0.1$ & $\bm{0.65} \pm 0.03$ & $0.85 \pm 0.02$ & $0.73 \pm 0.08$ & $0.74 \pm 0.04$  \\
\midrule
\multicolumn{3}{c}{Graphormer} & $0.68 \pm 0.02$ & $0.60 \pm 0.03$ & $0.60 \pm 0.14$ & $0.70 \pm 0.03$ & $0.63 \pm 0.01$ & $0.64 \pm 0.11$ & $0.73 \pm 0.03$ & $0.66 \pm 0.05$ & $0.65 \pm 0.18$\\
\multicolumn{3}{c}{Naive-Bayesian} & $0.50 \pm 0.02$ & $0.50 \pm 0.02$ & $0.54 \pm 0.03$ &  $0.50 \pm 0.02$ & $0.50 \pm 0.02$ & $0.55 \pm 0.02$ & $0.50 \pm 0.02$ & $0.50 \pm 0.01$ & $0.55 \pm 0.01$\\

\bottomrule
\bottomrule

\multicolumn{3}{c}{\multirow{2}{*}{\textbf{Dataset: Large}}} & \multicolumn{3}{c}{100}   &  \multicolumn{3}{c}{200} &  \multicolumn{3}{c}{1000} \\

& & & AUC   & Accuracy & F1 & AUC   & Accuracy & F1 & AUC   & Accuracy & F1\\
\midrule
\multirow{6}{*}{VN-Solver}     & \multirow{3}{*}{Gray}  &  Circular & $0.58 \pm 0.33$ & $0.45 \pm 0.02$ & $0.62 \pm 0.02$  &  $0.44 \pm 0.39$ & $0.44 \pm 0.02$ & $0.61 \pm 0.02$ & $0.96 \pm 0.03$ & $0.92 \pm 0.03$ & $0.92 \pm 0.04$\\
    &   &  Spiral & $0.51 \pm 0.19$ & $0.53 \pm 0.12$ & $0.5 \pm 0.28$  &  $0.72 \pm 0.24$ & $0.63 \pm 0.23$ & $0.72 \pm 0.15$ & $0.98 \pm 0.02$ & $0.95 \pm 0.02$ & $\bm{0.94} \pm 0.02$ \\
     &   &  Random & $0.51 \pm 0.17$ & $0.48 \pm 0.06$ & $0.37 \pm 0.34$  & $0.54 \pm 0.14$ & $0.57 \pm 0.09$ & $0.26 \pm 0.36$ & $0.81 \pm 0.02$ & $0.72 \pm 0.06$ & $0.72 \pm 0.03$ \\
         \cmidrule{2-12}
      & \multirow{3}{*}{\makecell{Uniform \\ color}}  &  Circular &  $0.83 \pm 0.08$ & $0.72 \pm 0.16$ & $\bm{0.74} \pm 0.10$ & $0.93 \pm 0.04$ & $0.90 \pm 0.07$ & $\bm{0.90} \pm 0.08$ & $0.98 \pm 0.01$ & $0.94 \pm 0.02$ & $\bm{0.94} \pm 0.03$\\
    &   &  Spiral &  $0.86 \pm 0.04$ & $0.78 \pm 0.03$ & $\bm{0.75} \pm 0.07$ & $0.91 \pm 0.11$ & $0.81 \pm 0.18$ & $\bm{0.83} \pm 0.12$ & $0.98 \pm 0.01$ & $0.95 \pm 0.01$ & $\bm{0.95} \pm 0.02$ \\
     &   &  Random & $0.53 \pm 0.03$ & $0.48 \pm 0.02$ & $0.51 \pm 0.29$ & $0.51 \pm 0.06$ & $0.47 \pm 0.00$ & $0.64 \pm 0.00$ & $0.47 \pm 0.08$ & $0.41 \pm 0.02$ & $0.58 \pm 0.01$\\
     \cmidrule{2-12}
          & \multirow{2}{*}{\makecell{Random \\ color}}  &  Circular & $0.59 \pm 0.18$ & $0.54 \pm 0.09$ & $0.63 \pm 0.03$ & $0.87 \pm 0.12$ & $0.77 \pm 0.17$ & $0.80 \pm 0.10$ & $0.97 \pm 0.01$ & $0.91 \pm 0.02$ & $0.91 \pm 0.03$\\
          & & Spiral & $0.77 \pm 0.13$ & $0.58 \pm 0.16$ & $0.64 \pm 0.09$ &  $0.82 \pm 0.27$ & $0.79 \pm 0.17$ & $0.81 \pm 0.10$ & $0.98 \pm 0.01$ & $0.94 \pm 0.02$ & $0.93 \pm 0.03$ \\
\midrule
\multicolumn{3}{c}{Graphormer} &  $0.83 \pm 0.01$ & $0.76 \pm 0.08$ & $\bm{0.74} \pm 0.12$ & $0.83 \pm 0.02$ & $0.83 \pm 0.02$ & $\bm{0.83} \pm 0.02$ & $0.93 \pm 0.01$ & $0.92 \pm 0.01$ & $0.92 \pm 0.01$\\
\multicolumn{3}{c}{Naive-Bayesian} & $0.49 \pm 0.02$ & $0.51 \pm 0.02$ & $0.51 \pm 0.03$ & $0.50 \pm 0.01$ & $0.50 \pm 0.01$ & $0.54 \pm 0.04$ & $0.5 \pm 0.02$ & $0.49 \pm 0.01$ & $0.51 \pm 0.03$\\

\bottomrule
\end{tabular}%
} 
\end{table*}

\section{Empirical Studies}
In this paper, we focus on the Hamiltonian cycle problem. Based on the House of Graphs \cite{coolsaet2023house}, we create two datasets, Small and Large. The Small dataset contains 4,115 graphs with 4 to 20 nodes, where 2,277 are Hamiltonian and 1,838 are non-Hamiltonian; the Large dataset consists of 13,192 graphs with 20 to 50 nodes, where 7,453 are Hamiltonian and 5,739 are non-Hamiltonian. We examine the following methods in our experiments.

\begin{itemize}
    \item 
    \textbf{VN-Solver}. Unless otherwise specified, we have $a/b=1$ for the circular layout and $r=0.3$ for the spiral layout; for all visualizations, each node occupies $2\times 2$ in pixels and the thickness of the line segments is 2 in pixels. We train ResNet using Adam with a learning rate of $0.001$ and an exponential learning rate decay of $0.09$. We set the maximum epoch as 200 and stop when the F1 score does not increase for 8 epochs.
    
    \item 
    \textbf{Graphormer}. We consider graph transformer \cite{ying2021transformers}, which is one of the state-of-the-art matrix-based methods. We adopt the graphormer-slim architecture composed of 12 attention layers and each layer has 8 heads. The training is done via Adam with a learning rate of $0.001$, and a decay of $0.01$. The early stopping is the same as that in the VN-Solver.
    
    \item 
    \textbf{Naive-Bayesian}. This is a feature-oblivious Bayesian method that makes the prediction based on the prior distribution over $\{0, 1\}$ estimated from the training samples. For example, the prediction is uniformly at random from $\{0, 1\}$, if half of the training samples are true instances.
    
\end{itemize}
For each learning-based method, the training-testing process is repeated five times, and we report the average performance in terms of common metrics together with the standard deviation. The testing size is fixed to be $500$; we experiment with sample sizes for training, where 80 (resp., 20) percent of the samples are used for training (resp., validation).

\subsection{Observations}

\begin{figure}[t]
\centering
\subfloat[{[circular, 100]}]{\label{fig: ellipse_80_20_all_seeds}\includegraphics[width=0.15\textwidth]
{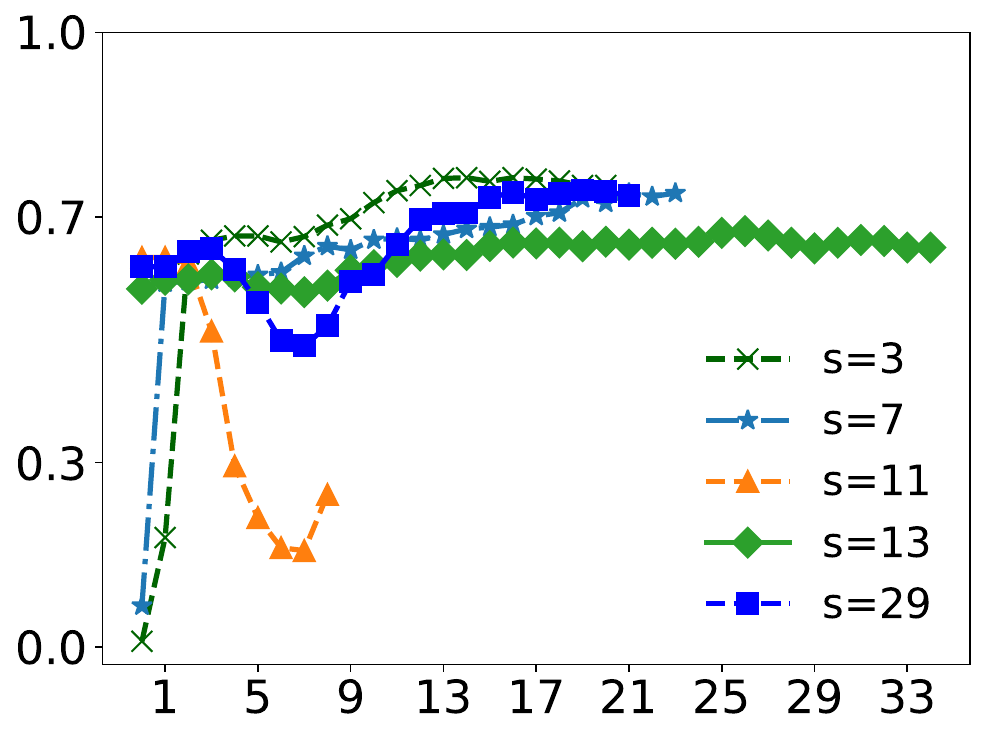}}\hspace{0mm}
\subfloat[{[spiral, 100]}]{\label{fig: spiral_80_20_all_seeds}\includegraphics[width=0.15\textwidth]{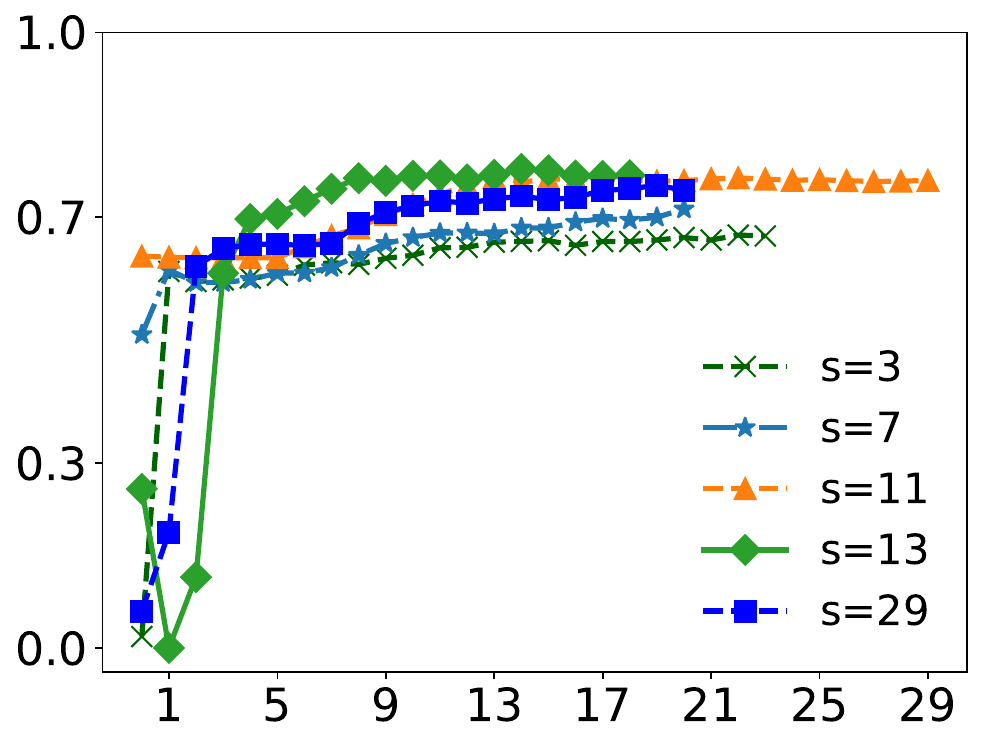}}\hspace{0mm}
\subfloat[{[random, 100]}]{\label{fig: random_80_20_all_seeds}\includegraphics[width=0.15\textwidth]{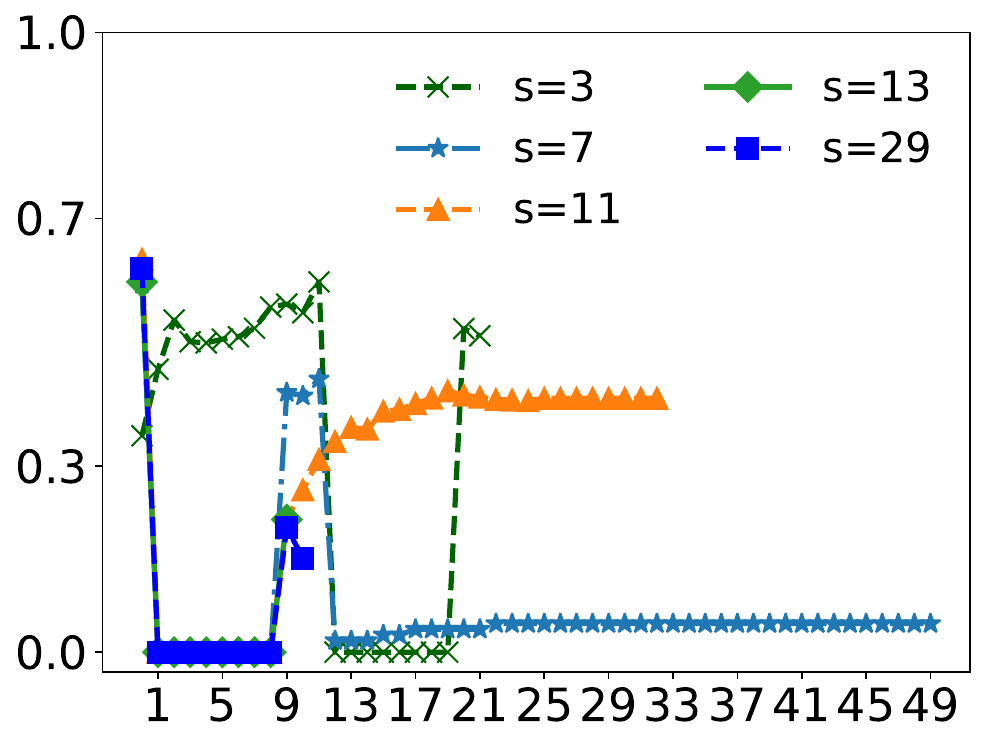}}\\

\subfloat[{[circular, 200]}]{\label{fig: ellipse_160_40_all_seeds}\includegraphics[width=0.15\textwidth]{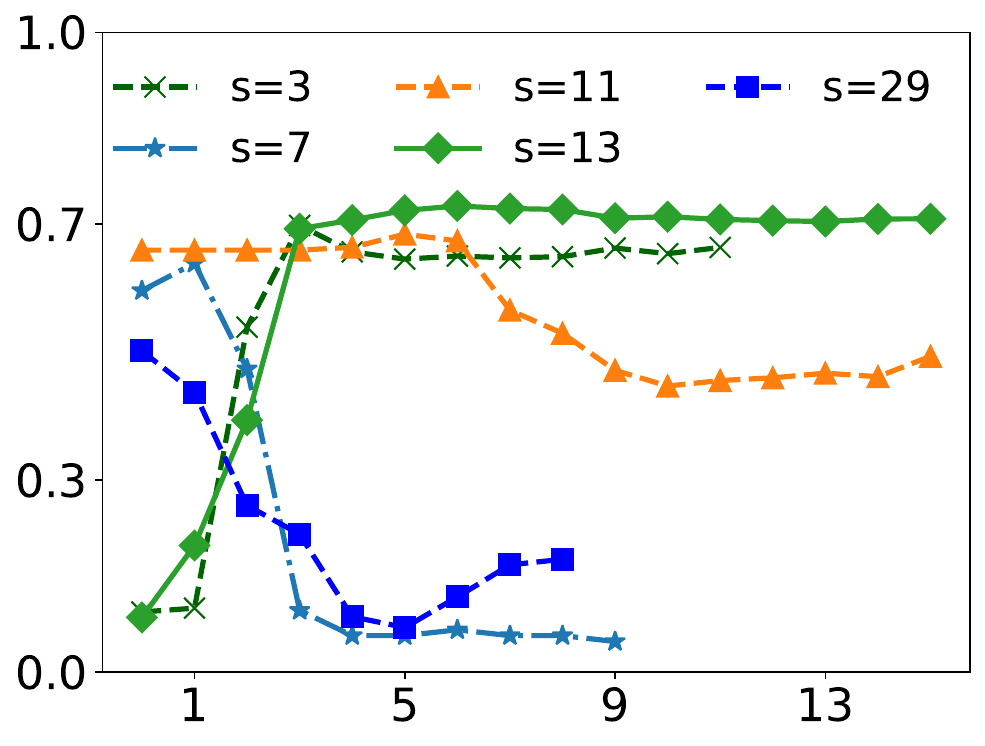}}\hspace{0mm}
\subfloat[{[spiral, 200]}]{\label{fig: ellipse_160_40_all_seeds}\includegraphics[width=0.15\textwidth]{images/ellipse_160_40_all_seeds.pdf}}\hspace{0mm}
\subfloat[{[random, 200]}]{\label{fig: random_160_40_all_seeds}\includegraphics[width=0.15\textwidth]{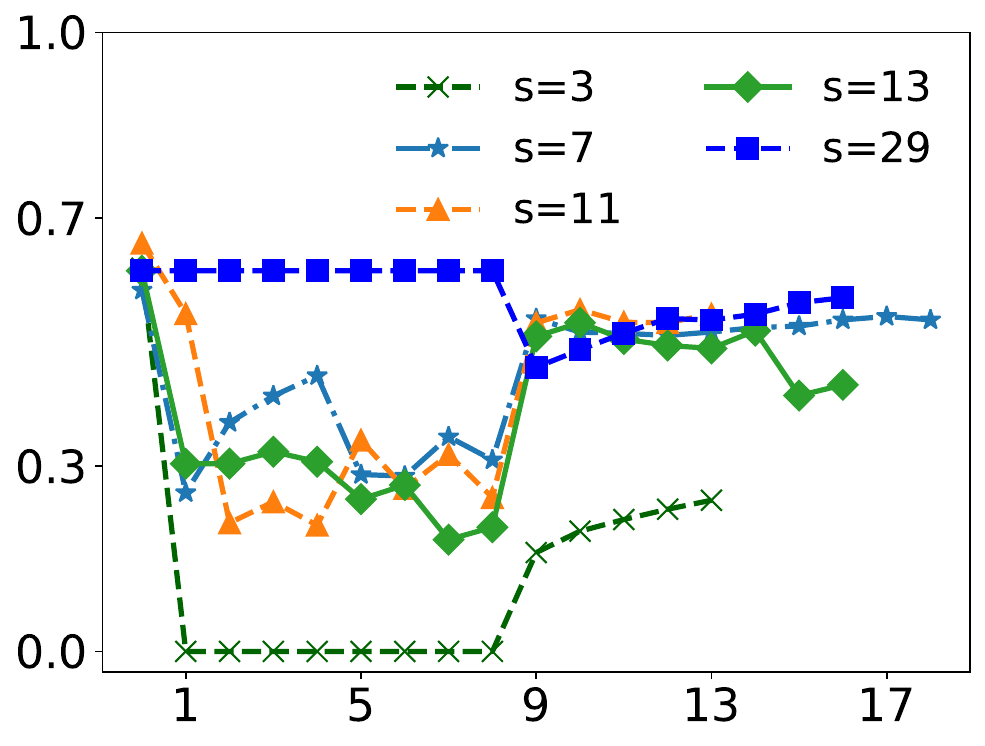}}

\caption{\small The F1 score of VN-Solver after each epoch on dataset Small. Each sub-float is labeled by the layout method and the sample size in training, and it shows the results under five different seed values. The early stopping ensures that a decent checkpoint (not necessarily the last epoch in the figure) is selected as the final model. }
\label{fig: epoch}
\vspace{-3mm}
\end{figure}

\textbf{Feasibility.} The main results can be found in Table \ref{table: results_all}. In addition, the generalization performance of VN-Solver after each training epoch is given in Fig. \ref{fig: epoch}, where results for circular and spiral layouts suggest a reasonable learning process, while the random visualizations are not very informative. From these results, it is clear that VN-Solver can perform better when more training data is given or more training epochs are used. This confirms that VN-Solver indeed can learn from data towards solving the Hamiltonian cycle problem, which gives the first piece of evidence supporting the feasibility of vision-based neural combinatorial solvers.

\textbf{Effectiveness.} Promisingly, the results in Table \ref{table: results_all} also suggest that VN-Solver is not only feasible but also no less effective than the state-of-the-art matrix-based method in most cases. In particular, it is comparable to Graphormer when the visualizations are gray, and it outperforms Graphormer under the uniform-color scheme. Notably, the results of VN-Solver and Graphormer are statistically significant, as they clearly outperform Naive-Bayesian.

\textbf{Coloring scheme.} One interesting observation is that colored visualization can consistently lead to better performance than gray ones do. This is especially true when the training size is not very large: with 200 training samples on the Large dataset, switching the coloring scheme from gray to uniform can increase the F1 score from 0.61 to 0.90 under the circular layout. One plausible reason is that the graph structure is more salient to image classifiers like ResNet when the nodes and edges are colored.  For similar reasons, we see that switching from uniform-color scheme to random-color scheme will slightly decrease the performance.  

\textbf{Other factors.} Table \ref{table: addition}(a) shows the results of VN-Solver under the circular layout with different node sizes and edge thicknesses. In general, we see that the performance becomes better when the nodes and edges are visualized with larger sizes. Of course, an extreme size can make the graph structure not recognizable -- for example, $(x,y)=(100,100)$ as shown in Fig. \ref{fig: c-1-1-100-100-g}. Finally, we observe that the performance is also sensitive to the parameters of the layouts, as shown in Table \ref{table: addition}(c) and \ref{table: addition}(d). For the circular layout, increasing the ratio tends to decrease the performance. For the spiral layout, visualizations with a small spiral factor $r$ can be better recognized by ResNet, which is intuitive since, for example, Fig. \ref{fig: s-01-05-01-g} appears to be more structured than Fig. \ref{fig: s-07-05-01-g}.

\begin{table}[pt!]
\renewcommand{\arraystretch}{1.2} 
\caption{\small Additional results of VN-Solver. The results in subtables (a)-(c) are F1 scores with 200 samples for training under the uniform-color scheme.}
\vspace{-3mm}
\label{table: addition}

{\small
\begin{tabular}{@{}  l@{\hspace{2mm}}  l@{\hspace{2mm}}  c@{\hspace{2mm}} c@{\hspace{2mm}} c@{\hspace{2mm}} c@{\hspace{2mm}} }
\multicolumn{6}{@{}p{0.45\textwidth}}{\textbf{Table \ref{table: addition}(a):} Results of the circular layout, where each cell $(x,y)$ means that each node (resp., edge) is scaled by a factor of $x$ (resp., $y$).} \\
\toprule
  & y=0.01 & y=0.1 & y=1 & y=10 & y=100\\
\midrule
x=0.01 &  $0.63 \pm 0.03$ & $0.7 \pm 0.04$ & $0.71 \pm 0.05$ & $0.73 \pm 0.04$ & $0.68 \pm 0.02$\\
x=0.5 &  $0.64 \pm 0.04$ & $0.69 \pm 0.04$ & $0.72 \pm 0.02$ & $0.74 \pm 0.04$ & $0.66 \pm 0.06$\\
x=5 & $0.64 \pm 0.06$ & $0.71 \pm 0.03$ & $0.74 \pm 0.03$ & $0.74 \pm 0.02$ & $0.66 \pm 0.06$\\
x=100 & $0.65 \pm 0.05$ & $0.72 \pm 0.03$ & $0.75 \pm 0.06$ & $0.77 \pm 0.02$ & $0.66 \pm 0.05$\\
\bottomrule
\bottomrule
\end{tabular}%
}

{\small
\begin{tabular}{@{}  l@{\hspace{8mm}}  c@{\hspace{5mm}}  c@{\hspace{5mm}} c@{\hspace{5mm}} c@{}}
\multicolumn{5}{@{}p{0.45\textwidth}}{\textbf{Table \ref{table: addition}(b):} Circular layout with different ratios of $a/b$.} \\
\toprule
  & $a/b=0.001$ & $a/b=0.01$ & $a/b=0.1$ & $a/b=1$ \\
\midrule
Small& $0.62 \pm 0.02$ & $0.63 \pm 0.13$ & $0.63 \pm 0.05$ & $0.69 \pm 0.04$\\
Large & $0.67\pm0.11$ & $0.77 \pm 0.15$ & $0.78 \pm 0.07$ & $0.90 \pm 0.08$\\
\bottomrule
\bottomrule
\end{tabular}%
} 

{\small
\begin{tabular}{@{}  l@{\hspace{8mm}}  c@{\hspace{5mm}}  c@{\hspace{5mm}} c@{\hspace{5mm}} c@{} }
\multicolumn{5}{@{}p{0.45\textwidth}}{\textbf{Table \ref{table: addition}(c):} Spiral layout with different $r$.} \\
\toprule
  & $r=0.1$ & $r=0.3$ & $r=0.5$ & $r=1$ \\
\midrule
Small & $0.73 \pm 0.02$ & $0.66 \pm 0.06$ & $0.66 \pm 0.02$ & $0.63 \pm 0.06$ \\
Large & $0.85 \pm 0.04$ & $0.89 \pm 0.02$ & $0.86 \pm 0.05$ & $0.87 \pm 0.05$\\
\bottomrule
\end{tabular}%
} 
\end{table}

\begin{figure}[t]
\centering
\subfloat[Embedding]{\label{fig: circular}\includegraphics[width=0.12\textwidth]
{images/c-1-1-300-10-r.pdf}}\hspace{5mm}
\subfloat[Style]{\label{fig: spiral_80_20_all_seeds}\includegraphics[width=0.12\textwidth, height=0.12\textwidth]{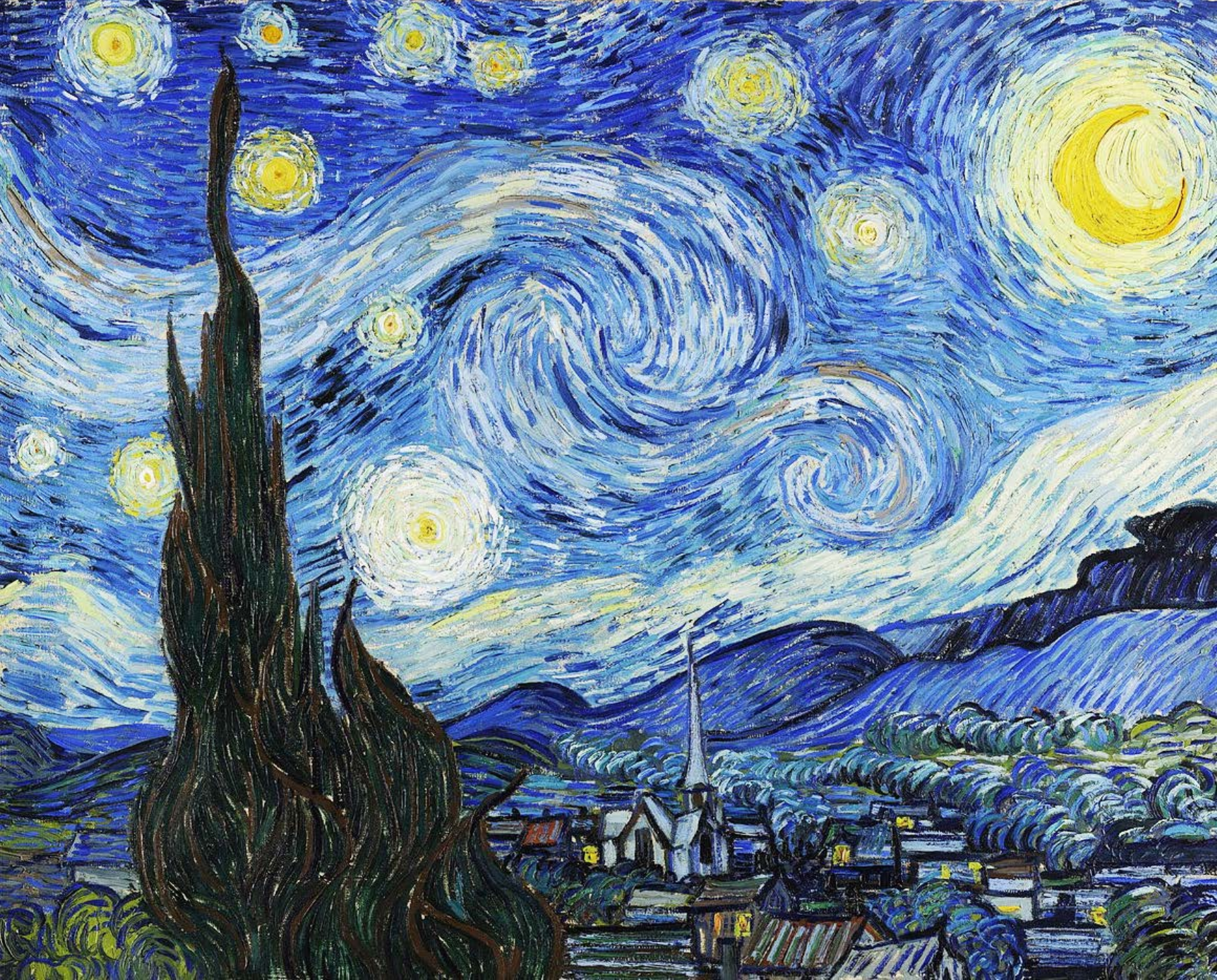}}\hspace{6mm}
\subfloat[Visualization]{\label{fig: random_80_20_all_seeds}\includegraphics[width=0.12\textwidth]{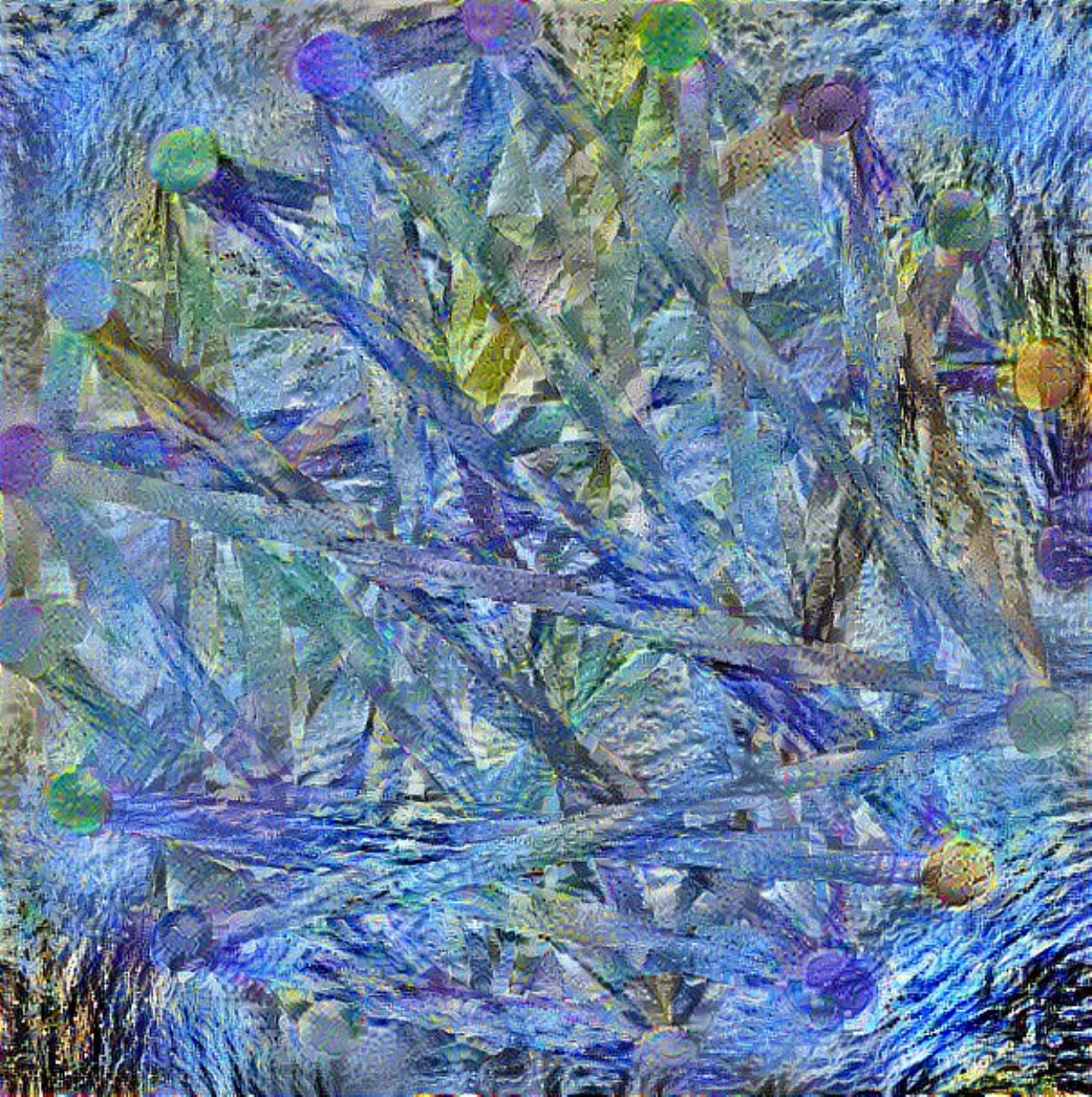}}\\

\caption{\small An example of generating coloring schemes (c) by giving an embedding (a) a style (b) using neural networks \cite{gatys2016image}. Is there a style that is useful in terms of deciding the Hamiltonian cycle?}
\label{fig: style}
\vspace{-3mm}
\end{figure}

\newpage

\section{Further Discussions}
This paper presents the framework of vision-based neural solvers, which are conceptually different from the common practice based on matrix processing. The experiments have confirmed the feasibility and effectiveness of such methods for the Hamiltonian cycle problem, with the hope to pave a new avenue for developing neural combinatorial solvers. We close our paper by listing the following research issues that we believe deserve future investigations in depth.
\begin{itemize}
    \item \textbf{Wider applicability.} It is of interest to examine the feasibility of vision-based methods for other deterministic graph optimization problems. For example, can neural solvers effectively decide graph isomorphism based on the visualizations? 
    \item \textbf{Advanced embedding methods.} What is the best graph embedding method with respect to solving a given graph optimization problem? This question is quite open and has to be discussed on a case-by-case basis. In addition to classic methods like circular and spiral layouts, one can even use generative models to design learnable embedding methods. In another issue, besides the 2D Euclidean space, one can embed graphs in any other metric space, which creates new research opportunities.

    \item \textbf{The best visualization style.} Following the investigations on the coloring scheme, node size, and edge thickness, we are wondering how to design better visualization styles. For example, while the random-color scheme does not exhibit extra benefits for our cases (i.e., the Hamiltonian cycle problem), images of perceptual styles with possible semantic meanings might be better recognized by machine learning models thereby offering better generalization performance. Immediately and somehow surprisingly, techniques for image style analysis \cite{karayev2013recognizing,gatys2016image} become a potential component of neural solvers for graph optimization (See Fig. \ref{fig: style} for an example).
\end{itemize}


\bibliographystyle{ACM-Reference-Format}
\bibliography{acmart}

\end{document}